Chapter 1

# NEURAL-NETWORK TECHNIQUES FOR VISUAL MINING CLINICAL ELECTROENCEPHALOGRAMS

Vitaly Schetinin[1], Joachim Schult[2], and Anatoly Brazhnikov[3]
[1]*University of Exeter, UK;* [2]*University of Jena, Germany;* [3]*Ansvazer Consulting, Canada*

**Abstract**: In this chapter we describe new neural-network techniques developed for visual mining clinical electroencephalograms (EEGs), the weak electrical potentials invoked by brain activity. These techniques exploit fruitful ideas of Group Method of Data Handling (GMDH). Section 2 briefly describes the standard neural-network techniques which are able to learn well-suited classification modes from data presented by relevant features. Section 3 introduces an evolving cascade neural network technique which adds new input nodes as well as new neurons to the network while the training error decreases. This algorithm is applied to recognize artifacts in the clinical EEGs. Section 4 presents the GMDH-type polynomial networks learnt from data. We applied this technique to distinguish the EEGs recorded from an Alzheimer and a healthy patient as well as recognize EEG artifacts. Section 5 describes the new neural-network technique developed to induce multi-class concepts from data. We used this technique for inducing a 16-class concept from the large-scale clinical EEG data. Finally we discuss perspectives of applying the neural-network techniques to clinical EEGs

**Key words**: Classification model, pattern visualization, neural network, cascade architecture, feature selection, polynomial, decision tree, electroencephalogram

## 1. INTRODUCTION

Data mining as a process of discovering interesting patterns and relations in data presented by *labeled examples* can be referred to inducing *classification models* or *classifiers* which assign an unknown example to one of the given classes with acceptable accuracy. A typical classification problem is presented by a *data set* of labeled examples which are characterized by *variables* or *features*. Experts assume such features which make the distinct contribution to the classification problem. Such features are called *relevant*. However among these features may be assumed *irrelevant* and/or *redundant*: the first can seriously hurt the classification accuracy whereas the second are useless for the classification and can obstruct understanding how decisions are arrived at. Both the irrelevant and redundant features have to be discarded.

Solving classification problem user has to induce or learn a classification model from the *training data* set and test its performance on the *testing data* set of the labeled examples. These data must be disjoint in order to objectively evaluate how well the classification model can classify unseen examples.

Besides that users, such as medical experts, need not only to classify unseen examples but also verify decisions by analyzing the underlying causal relations between the involved features and the model outcome. Such an analysis can be comprehensively done by visualizing a discovered model and/or discovered patterns (Kovalerchuk & Vityaev; 2000). Some data mining methods can provide the visualization of classification model as well as patterns. For example, we may visualize an induced decision tree model where each branch visualizes an individual pattern. However using neural-network techniques we can visualize an induced network but we cannot visualize the interesting patterns. This issue is very critical for medical experts who need to interpret data mining results in a visual form additionally to textual description.

In this chapter we describe new neural-network techniques developed to provide both the visualization of classification models and the visualization of patterns. The advantages of these techniques are illustrated by mining medical data such as electroencephalograms (EEGs), the weak electrical potentials invoked by brain activity, whose spectral characteristics are taken as visual features. Within this chapter we compare some data mining techniques and the new techniques in the respect to the above aspects of visualization. The achieved results show that in addition to textual presentation EEG-experts can visually interpret discovered classification models and patterns.

Applying data mining techniques EEG-experts often cannot properly assume relevant features and avoid irrelevant and redundant. Besides that, some features become relevant being taken in account in the combination with others. In such cases data mining techniques exploit a special learning strategy capable of *selecting* relevant features during the induction of classification model (Duda & Hart 2000; Farlow 1984; Madala & Ivakhnenko 1994; Müller & Lemke 2003). Such a strategy allows experts to learn classification models more accurately than strategies selecting features before learning.

Surveying data mining methods we see that most of them aimed to extract comprehensible models imply a *trade-off* between classification accuracy and representation complexity (Avilo Garcez, Broda & Gabbay 2001; Setiono 2000, Towell & Shavlik 1993). Less work has been undertaken to study on the methods capable of discovering the comprehensible models without decreasing their classification accuracy.

Below we describe new neural-network techniques developed for visual mining clinical EEGs. Exploiting fruitful ideas of Group Method of Data Handling (GMDH) of Ivakhnenko (Madala & Ivakhnenko 1994; Müller & Lemke 2003) these techniques are able to induce the comprehensible classification models and meantime keep their classification error down.

In section 2 we briefly describe the standard neural-network techniques, including cascade-correlation architecture, which are able to learn well-suited classification modes from data. These methods however cannot generalize well in the presence of irrelevant and/or noise features.

Section 3 introduces an evolving cascade neural network technique, which adds new input nodes as well as new neurons to the network while the training error decreases. The resultant networks have a nearly minimal number of input variables and hidden neurons that allow classifying new examples well. We applied this algorithm to recognize artifacts in the clinical EEGs.

Section 4 presents the GMDH-type polynomial networks learnt from data. These networks are represented as concise sets of short-term polynomials and can be presented in visual form. Moreover, the GMDH-type neural networks can generalize even better than the standard fully connected neural networks. We applied this technique to distinguish the EEGs recorded from an Alzheimer and a healthy patient as well as recognize EEG artifacts.

Section 5 describes the new decision tree neural-network technique developed to induce multi-class concepts from data. We used this technique for inducing a 16-class concept from the large-scale clinical EEG data recorded from sleeping newborns. This concept assists clinicians to predict some brain development pathologies of newborns. Finally we discuss perspectives of applying the neural-network techniques to clinical EEGs

## 2. NEURAL NETWORK BASED TECHNIQUES

In this section we briefly describe a standard technique used in our experiments for training feed-forward neural networks (FNN) by a back-propagation algorithm. Then we describe the cascade-correlation architecture and finally discuss the shortcomings and advantages of these techniques.

### 2.1    A Standard Neural-Network Technique

A standard neural-network technique exploits a *feed-forward* fully connected network consisting of the *input nodes*, *hidden* and *output* neurons which are connected each other by the adjustable *synaptic weights* (Bishop 1995). This technique implies that a *structure* of neural network has to be predefined properly. This means that users must preset an appropriate number of the input nodes and hidden neurons and apply a suitable *activation function*. For example, the user may apply a *sigmoid* activation function described as

$$y = f(\mathbf{x}, \mathbf{w}) = 1/(1 + \exp(-w_0 - \Sigma_i^m w_i x_i)), \tag{1}$$

where $\mathbf{x} = (x_1, ..., x_m)^T$ is a $m \times 1$ input vector, $\mathbf{w} = (w_1, ..., w_m)^T$ is a $m \times 1$ synaptic weight vector, $w_0$ is a bias term and $m$ is the number of input variables.

Then the user has to select a suitable learning algorithm and then properly set its parameters such as the *learning rate* and the number of the *training epochs*. Note that when the neural networks include at least two hidden neurons, the learning algorithms with *error back-propagation* usually provide the best performance in the term of the classification accuracy (Bishop 1995).

Within the standard technique first the learning algorithm initializes the synaptic weights $\mathbf{w}$. The values of w are updated while the *training error* decreases for a given number of the training epochs. The resultant classification error is dependant on the given learning parameters as well as on the initial values $\mathbf{w}^0$ of neuron weights. For these reasons neural networks are trained several times with random values of initial weights and different learning parameters. This allows the user to avoid local minima and find out a neural network with a near minimal classification error.

After training the user expects that the neural network can classify new inputs well and its classification accuracy is acceptable. However the learning algorithm may fit the neuron weights to specifics of training data, which are absent in new data. In this case neural networks become to be *over-fitted* and cannot generalize well. Within the standard technique the generalization ability of the trained network is evaluated on a *validation* subset of the labeled examples, which have not been used for training the network.

Figure 1 depicts a case when after $k^*$ training epochs the validation error starts to increase while the training error continues to decrease. This means that after $k^*$ training epochs the neural network becomes over-fitted. To prevent over-fitting, we can update the neuron weights while the validation error decreases.

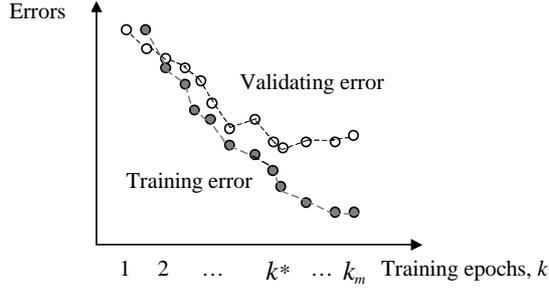

Figure 1: Learning curves for the training and validating sets.

When classification problems are characterized in the *m*-dimensional space of input variables, the performance of neural networks may be radically improved by applying the Principal Component Analysis (PCA) to training data (Bishop 1995). The PCA may significantly reduce the number of the input variables and consequently the number of synaptic weights, which are updated during learning. A basic idea behind the PCA is to turn the initial variables so that the classification problem might be resolved in a reduced input space.

Figure 2 depicts an example of a classification problem resolved in a two-dimensional space of the input variables $x_1$ and $x_2$ by using a separating function $f_1(x_1, x_2)$. However we can turn the $x_1$ and $x_2$ so that this problem might be solved in one-dimensional input space of a principal component $z_1 = a_1 x_1 + a_2 x_2$, where $a_1$ and $a_2$ are the coefficients of a linear transformation. In this case a new separating function is $f_2(z_1)$ that is equal to 0 if $z_1 < \vartheta_1$ and 1 if $z_1 \geq \vartheta_1$, where $\vartheta_1$ is a threshold learnt from the training data represented by the new variable $z_1$.

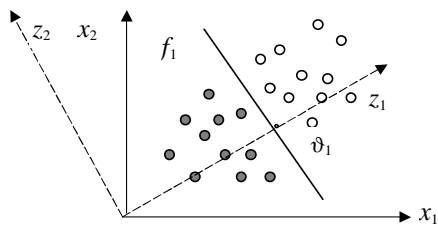

Figure 2: An example of two-dimensional classification problem.

As we see, the new components $z_1$ and $z_2$ make the different contribution to the variance of the training data: the first component contributes much greater than the second. This example demonstrates how the PCA can rationally reduce the input space. However users using PCA must properly define a variance level and the number of components making the contribution to the classification.

Thus, using the standard technique, we may find out a suitable neural-network structure and then fit its weights to the training data while the validation error decreases. Each neural network with a given number of input nodes and hidden neurons should be trained several times, say 100 times.

Thus, we can see that the standard technique is computationally expensive. For this reason, users use fast learning algorithms, for instance, a back-propagation algorithm by Levenberg-Marquardt (Bishop 1995).

## 2.2    A Cascade-Correlation Architecture

To solve classification and pattern recognition problems, Fahlman & Lebiere (1990) proposed a *cascade-correlation architecture* of neural networks. The neural networks with the cascade-correlation architecture differ from the above networks with a predefined structure. In contrast to the last, cascade networks start learning with only one neuron. Then the algorithm adds and trains new neurons creating a multi-layer structure. The new neurons are added to the networks as long as the residual classification error decreases. Thus, the cascade-correlation learning algorithm allows growing neural networks of a *near optimal size* required to generalize well (Farlow 1984, Iba, deGaris & Sato 1994; Madala & Ivakhnenko 1994; Müller & Lemke 2003).

Figure 3 depicts an example of cascade-correlation architecture consisting of four input nodes $x_1, \ldots, x_4$, two hidden neurons $z_1$ and $z_2$, and one output neuron $y$. The first hidden neuron is connected to all the input nodes, and the output neuron is connected to all the input nodes as well as to the hidden neurons $z_1$ and $z_2$.

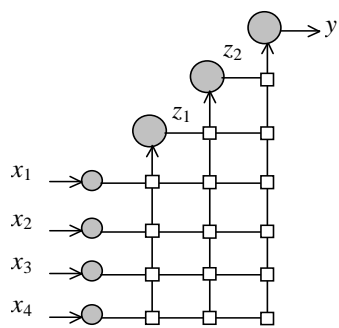

Figure 3: An example of cascade-correlation architecture.

The learning of cascade-correlation architecture is based on the following ideas. The first is to build up the cascade architecture by adding new neurons connected to all the input nodes and previous hidden neurons. The second is that the learning algorithm attempts to reduce the residual error by updating weights of the new neuron, that is, each time only the output neuron is trained. The third idea is to add one-by-one new neurons to the network while its residual error decreases.

The main steps of the learning cascade-correlation algorithm are described below.

```
nnet = []; % initializing
error = n;
new-error = error - 1;
while new-error < error
  nnet = add-new-neuron(nnet);
  nnet = train-neuron(nnet, X, Y);
  new-error = calc-error(net(nnet, X) - Y);
  error = new-error;
end
nnet = cut(nnet);
```

Here *n*, X, and Y are the number of training examples, the input data and a target vector, respectively. The procedure cut excludes the last neuron from the trained cascade network and then returns the result to the nnet.

There are two advantages of the cascade neural networks. First, no size and connectivity of neural networks are predefined, that is, the network is automatically built up. Second, the cascade network learns fast because each of its neurons is trained independently from other neurons.

However, the algorithm can train cascade network well if all the input variables are relevant to the classification problem. In section 3 we will describe a new algorithm, which can train cascade neural networks in the presence of irrelevant features.

## 3. EVOLVING CASCADE NEURAL NETWORKS

In this section we describe an evolving cascade neural network technique, which adds new input nodes as well as new neurons to the network while the training error is decreased. This algorithm is used to recognize artifacts in the clinical EEGs.

### 3.1 An Evolving Cascade Neural Network

Let us assume a classification problem presented by $m$ input variables $x_1$, ..., $x_m$ some of which may be irrelevant or noisy. In this case a standard pre-processing technique used for selecting relevant variables may fail because this technique does not consider useful combinations of the input variables. More suitable strategy is to select relevant features during learning. To do so let us define the neurons in which the number of inputs, $p$, increases as follows $p = r + 1$, where $r = 0, 1, 2, ...$ is the number of layer in the cascade network. So for $r = 0$ there are $m$ neurons with one input variable. For the first layer there are neurons with $p = 2$ inputs and so on.

Let us now fit all the neurons for $r = 0$. Then among these neurons can be found out one that provides the best performance on the validation data set. Fix an input variable of this neuron, $x_{i1}$, in order to connect it with all the neurons that will be added to the network.

At the first layer the algorithm trains the candidate-neurons with two inputs: the variable $x_{i1}$ and one of the remaining input variables. The neuron with the best performance is added to the network.

Each following neuron is connected with the variable $x_{i1}$ and the outputs of all the previous neurons. For the second layer the candidate-neurons have three inputs: the first is connected with the output of the previous neuron, the second input with the input $x_{i1}$, and the third with one of the input variables $x_1, ..., x_m$.

Defining a sigmoid activation function of the neurons, we can write the output $z_r$ for the $r$th neuron as follows:

$$z_r = f(\mathbf{u}, \mathbf{w}) = 1/(1 + \exp(-w_0 - \Sigma_i^p u_i w_i)), \qquad (2)$$

where $\mathbf{u} = (u_1, ..., u_p)$ is a $p \times 1$ input vector of the $r$th neuron, $\mathbf{w} = (w_1, ..., w_m)$ is a $m \times 1$ vector of synaptic weights, and $w_1$ is a bias term.

The idea behind our learning method is that the relevance of an input variable connected to the candidate-neuron can be estimated in *ad hoc* manner. The learning algorithm starts to train the candidate-neurons with one input variable and then step-by-step add the new inputs and new neurons to the network. As a result the internal connections of the cascade neural network are built up accordingly to the best performance achievable in each new layer. So building the cascade network our algorithm exploits a greedy search heuristic.

Within our technique, the performance of the network, $C_r$, is evaluated for each candidate-neuron at the $r$th layer. The value of $C_r$ is dependent on the generalization ability of the trained candidate-neuron with the given connections. Clearly, the neuron connected to the irrelevant connections cannot properly classify the validating examples and subsequently its value of $C_r$ is high.

If value $C_r$ calculated for the $r$th neuron is less than value $C_{r-1}$ calculated for the previous $(r-1)$th neuron, the connections selected for the $r$th neuron are relevant, otherwise they are irrelevant. Formally this heuristic can be described by the following inequality:

*if* $C_r < C_{r-1}$, *then* the connections are relevant,  (3)
 *else* irrelevant.

If inequality (3) is met, the $r$th neuron is added to the network. If no neurons satisfy this inequality, the algorithm stops. As a result an $r$th neuron providing a minimal validation error is assigned to be an output neuron for the cascade network.

## 3.2   An Algorithm for Evolving Cascade Neural Networks

Adding new features and neurons as they are required the cascade neural network is evolved during learning. The main steps of the evolving algorithm are described below.

```
X = [x_1, ..., x_m]; % a pool of m input variables
P = 1;      % the number of neuron inputs
% Train single-input neurons and calculate errors
for i = 1:m
  N1 = create-neuron(p, X(i));
  N1 = fit-weight(N1);
  E(i) = calc-error(N1);
end
[E1,F] = sort-ascend(E);
h = 1;     % the position of the variable in F
C0 = E1(h);
% Create a cascade network NN
```

```
NN = [];
r = 0;      % the number of hidden neurons
p = 2;
while h < m
 h := h + 1;
 V = [X(F(1)), X(F(h))];
 % Add links to the hidden neurons
 for j = 1:r
  V = [V, NN(j)];
 end
 % Create a candidate-neuron N1
 N1 = create-neuron(p, V];
 N1 = fit-weight(N1);
 C1 = calc-error(N1);
 if C1 < C0
  r := r + 1;
  p := r + 2;
  NN(r) = add-neuron(N1);
 end
end
```

The algorithm starts to learn the candidate-neurons with one input and then saves their validating errors in a pool E. The procedure `sort-ascend` arranges the pool E in an ascending order and saves the indexes of the input variables in a pool F. The first component of the F is an index of the input variable providing a minimal classification error $C_0$.

At the following steps the algorithm adds new features as well as new neurons to the network while the validation error $C_1$ calculated for the candidate-neuron $N_1$ decreases. The weights of candidate-neurons are updated until condition (3) is satisfied.

As a result the cascade neural network consisting of the *r* neurons is placed in the pool NN. The size of this network is nearly minimal because the stopping rule is met for a minimal number of neurons.

Below we describe an application of this algorithm to recognizing artifacts in clinical EEGs. These EEGs are characterized by numerous features some of which are noise, redundant or irrelevant to the classification problem.

## 3.3    An Evolving Cascade Neural Network

In our experiment we used the clinical EEGs recorded via the standard EEG channels C2 and C4 from two newborn during sleeping hours. Following Breidbach, Holthausen, Scheidt & Frenzel (1998) these EEGs were represented by spectral features calculated in 10-second segments for 6 frequency bands: subdelta (0-1.5 Hz), delta (1.5-3.5 Hz), theta (3.5-7.5 Hz), alpha (7.5-13.5 Hz), beta 1 (13.5-19.5 Hz), and beta 2 (19.5-25 Hz). Additionally for each band the values of relative powers and their variances were calculated for channels C3 and C4 and their sum, C3+C4. So the total number of the features was 72. Values of these features were normalized to be with zero mean and unit variance.

The normal segments and artifacts in the EEGs were manually labeled by an EEG-viewer which analyzed muscle and cardiac activities of patients recorded from additional channels. As an example of normal segments and artifacts, Figure 4 depicts the fragment of EEG containing 500 segments presented by 36 features. In this fragment the EEG-expert recognized segments 15, 22, 24, 84, and 85 as artifacts and the remaining as normal.

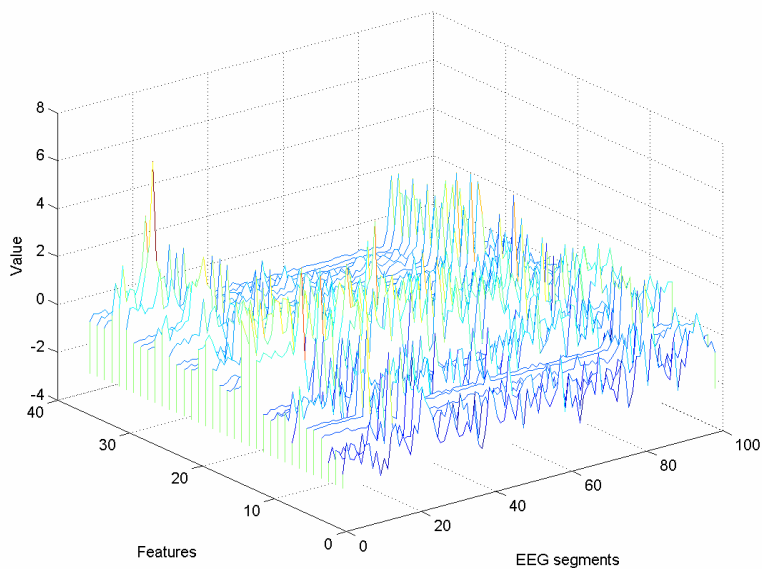

Figure 4: Fragment of EEG containing 100 segments presented by 36 features in which the EEG-viewer recognized 5 artifacts.

The patterns of EEG artifacts and normal segments can be visualized in a space of two principal components as depicted in Figure 5. Here artifacts and normal segments marked by the stars and the points, respectively.

Observing these patterns we see that the artifacts are located far away from the normal segments and therefore the statistical characteristics of these patterns should be different. The labeled EEG segments were merged and divided into the training and testing subsets containing 2244 and 1210 randomly selected segments including 209 and 99 artifacts, respectively. One-third of the training data we used for validation and two-third for fitting the neuron weights.

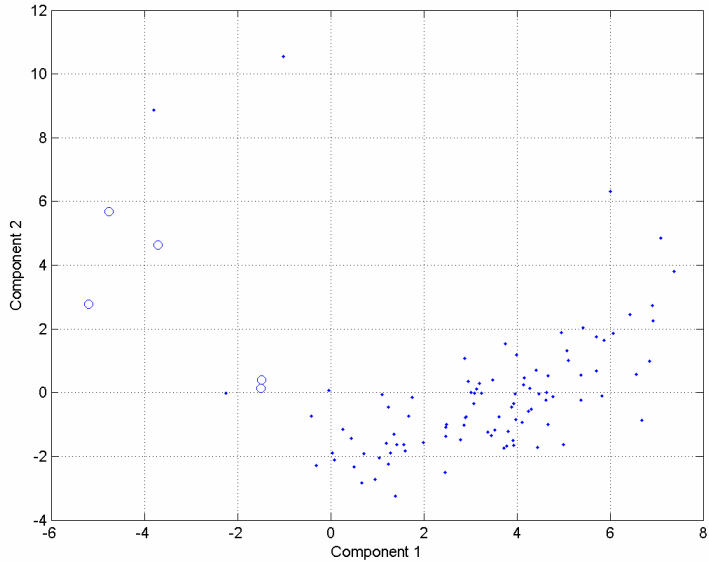

Figure 5: Pattern of EEG artifacts and normal segments in a space of two principal components. Here artifacts and normal segments are marked by the circles and points respectively.

Having trained 100 evolving cascade networks we selected one which provides a minimal training error equal to 3.92%. This network misclassified 3.31% out of the testing examples.

Figure 6 depicts a structure of this network containing four input nodes, three hidden neurons and one output neuron. From the given 72 initial features, the training algorithm has selected only four features *AbsPowBeta2*, *AbsPowAlphaC4*, *AbsPowDeltaC3*, and *AbsVarDelta* which are the absolute power of beta2 summed over C3 and C3, the absolute power of alpha in C4, the absolute power of delta in C3, and the absolute variance of delta summed over C3 and C4, respectively.

The inputs of the first neuron are connected to *AbsPowBeta2* and *AbsPowAlphaC4*. The output neuron is connected to *AbsPowBeta2*,

*1. Neural-Network Techniques for Visual Mining Clinical Electroencephalograms*     13

*AbsPowAlphaC4* and *AbsVarDelta* as well as to the outputs of the hidden neurons, the hidden variables, $z_1$, $z_2$, and $z_3$.

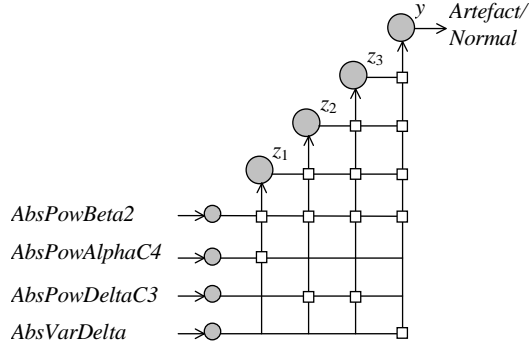

Figure 6: A cascade neural network learnt for recognizing artifacts and normal segments in clinical EEGs. The squares mean the synaptic connection.

    EEG-expert observing this model can conclude the following. First there are four features which make the most important contribution to the classification. These features are involved in the order of their significance – we can see that the most important feature is *AbsPowBeta2* and the less important is *AbsVarDelta*t. So the most important contribution to the artifact recognition in EEG of sleeping newborns is made by *AbsPowBeta2* which is calculated for a high frequency band. This fact directly corresponds to a rule used for recognizing muscle artifacts in sleep EEG of adults (Brunner, Vasko, Detka, Monahan, Reynolds, & Kupfer 1996).

    Second the discovered model shows the combinations between the selected features and hidden variables in the order of their classification accuracy. The EEG-expert can see that the maximal gain in the accuracy is achieved if the feature *AbsPowAlphaC4* is combined with *AbsPowBeta2*. The further improvement is achieved by combining the hidden variable $z_1$, which is a function of the above two features, and the new feature *AbsPowDeltaC3*. So the EEG expert can see the four combinations of the selected features and hidden variables $z_1$, …, $z_3$ listed in the order of increasing their classification accuracy, $p_1 < … < p_4$, as follows

    $z_1$: *AbsPowBet2* & *AbsPowAlphaC4* → $p_1$,
    $z_2$: $z_1$ & *AbsPowBet2* & *AbsPowDeltaC3* → $p_2$,
    $z_3$: $z_2$ & $z_1$ & *AbsPowBet2* & *AbsPowDeltaC3* → $p_3$,
    $z_4$: $z_3$ & $z_2$ & $z_1$ & *AbsPowBet2* & *AbsPowDelta* → $p_4$,

where $z_4 = y$ is the outcome of the classification model.

The third useful issue is that the synaptic connections in the discovered model are characterized by the real-valued coefficients which can be interpreted as the strength of relations between features and hidden variables. The large value of the coefficient, the stronger relation between the feature and hidden variable is.

In general such models can assist EEG-experts to present the underlying casual relations between the features and outcomes in a visual form. The visualization of the discovered models can be useful for understanding the nature of EEG artifacts.

In our experiments we compared the performance of the above classification model and the FNN trained on the same data. Using a sigmoid activation function and a standard neural-network technique, we found out that a FNN with four hidden neurons and 11 input nodes provides a minimal training error. The training and testing errors were equal to 2.97% and 5.54%, respectively.

Comparing the performances we conclude that the discovered cascade network slightly outperforms the FNN on the testing EEG data. The better performance is achieved because the cascade network is gradually built up by adding new hidden neurons and new connections. Each new neuron in the cascade network makes the most significant contribution to the artifact recognition among the all-possible combinations of the allowed number of features. This allows avoiding the contribution of the noise features and discovering most significant relations which can be visualized.

In this experiment the FNN has misclassified more testing examples than the classification model described above. Therefore we conclude that our cascade neural-network technique can more successfully recognize artifacts in clinical EEGs. In the same time the discovered classification model allows EEG-experts to present the basic relations between features and outcomes in visual form.

## 4. GMDH-TYPE NEURAL NETWORKS

In this section we describe GMDH-type algorithms, which allow inducing polynomial neural networks from data. The induced networks can generalize well because their size or complexity is near minimal. The induced networks are comprehensively described by concise sets of short-term polynomials, which are comprehensible for medical experts.

## 4.1 A GMDH Technique

GMDH-type neural networks are the multi-layered feed-forward networks consisting of the so-called *supporting neurons* (Farlow 1984, Madala & Ivakhnenko 1994; Müller & Lemke 2003). The supporting neurons have at least two inputs $v_1$ and $v_2$. A transfer function $g$ of these neurons may be described by short-term polynomials, for example, by a linear or non-linear polynomial:

$$y = g(v_1, v_2) = w_0 + w_1v_1 + w_2v_2, \qquad (4)$$

$$y = g(v_1, v_2) = w_0 + w_1v_1 + w_2v_2 + w_3v_1 v_2, \qquad (5)$$

where $w_0, w_1, w_2, \ldots$ are the polynomial coefficients or synaptic weights of the supporting neuron.

An idea behind GMDH-type algorithms is based on an evolution principle, which implies the *generation* and *selection* of the *candidate-neurons*. In the first layer, the neurons are connected to the input nodes, and in the second layer they are connected to the previous neurons selected. For selecting the candidate-neurons, which provide the best classification accuracy, GMDH exploits the *exterior* criteria which are capable of evaluating the generalization ability of neurons on the validation data set.

The user must properly define the number $F$ of the selected neurons providing the best classification accuracy. For example, the GMDH algorithm may combine the $m$ input variables by 2 in order to generate the first, $r = 1$, layer of candidate-neurons $y_1^{(1)}, \ldots, y_{L1}^{(1)}$, where $L_1 = \binom{m}{2} = m(m-1)/2$ is the number of the neurons. The algorithm trains these candidate-neurons and then selects $F$ best of them in order to generate the next layer. For generating the second layer, it is combined the outputs $y_1^{(1)}, \ldots, y_F^{(1)}$ of these $F$ selected neurons. The best performance of the algorithm it is achieved for $F = 0.4L_1$ (Farlow 1984, Madala & Ivakhnenko 1994).

In Figure 7 we depict an example of three-layer GMDH-type network.

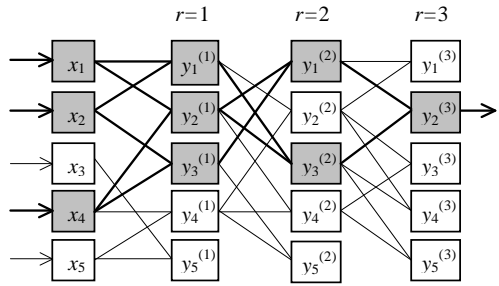

Figure 7: The structure of neural network grown by GMDH algorithm.

The neuron-candidates that were selected at the layers are depicted here as the gray boxes. A neuron $y_2^{(3)}$ that provides the best classification accuracy assigns to be an output neuron. A resulting polynomial network, as we can see, is the three-layer network consisting of 6 neurons and 3 input nodes. This network is described by a set of the following polynomials:

$$y_1^{(1)} = g_1(x_1, x_2),$$
$$y_2^{(1)} = g_2(x_1, x_4),$$
$$y_3^{(1)} = g_3(x_2, x_4),$$
$$y_2^{(3)} = g_4(y_1^{(2)}, y_3^{(2)}),$$
$$y_1^{(2)} = g_5(y_2^{(1)}, y_3^{(1)}),$$
$$y_3^{(2)} = g_6(y_1^{(1)}, y_2^{(1)}).$$

Thus, for the *k*th training example, we can calculate the output *y* of the neuron as

$$y = g(\mathbf{w}, \mathbf{v}^{(k)}), \ k = 1, \ldots, n,$$

where **w** is a weight vector, **v** is an input vector and *n* is the number of training examples.

For selecting *F* best neurons, the exterior criterion is calculated on the unseen examples of the validation set that have not been used for fitting the weights **w** of neurons. These examples are reserved by dividing the dataset **D** into two non-intersecting subsets $\mathbf{D}_A = (\mathbf{X}_A, \mathbf{y}_A^o)$ and $\mathbf{D}_B = (\mathbf{X}_B, \mathbf{y}_B^o)$, the training and validating data sets, respectively. The sizes $n_A$ and $n_B$ of these subsets is usually recommended to define $n_A \approx n_B$, i.e. $n_A + n_B = n$.

Let now find a weight vector **w**\* that minimizes the sum square error *e* of the neuron calculated on the subset $\mathbf{D}_A$:

$$e = \Sigma_k (g(\mathbf{v}^{(k)}, \mathbf{w}) - y^o_k)^2, \ k = 1, \ldots, n_A.$$

To obtain the desirable vector **w**\*, the conventional GMDH fits the neuron weights to the subset $\mathbf{D}_A$ by using a Least Square Method (LSM) (Bishop 1995 Farlow 1984, Madala & Ivakhnenko 1994), which can produce effective evaluations of weights under Gaussian distributed noise in the data. As noise in real-world data is often non-Gaussian (Duda & Hart 2000; Tempo, Calafiore & Dabbene 2003), we will use the learning algorithm described in Section 3, which does not require hypothesizing the noise structure.

Having found a desirable weight vector **w**\* on the subset $\mathbf{D}_A$, we can calculate the value $CR_i$ of the exterior criterion on the validation subset $\mathbf{D}_B$:

$$CR_i = \Sigma_k (g_i(\mathbf{v}^{(k)}, \mathbf{w}^*) - y^0_k)^2, \ k = 1, \ldots, n_B, \ i = 1, \ldots, L_r. \qquad (6)$$

We can see that the calculated value of $CR_i$ depends on the behavior of the *i*th neuron on the unseen examples of the subset $\mathbf{D}_B$. Therefore we may expect that the value of *CR* calculated on the data **D** would be high for the neurons with poor generalization ability.

The values $CR_i$ calculated for all the candidate-neurons at the $r$th layer are arranged in an ascending order:

$$CR_{i1} \leq CR_{i2} \leq \ldots \leq CR_{iF} \leq \ldots \leq CR_{iL},$$

so that the first $F$ neurons provide the best classification accuracy.

For each layer $r$, it is found out a minimal value $CR_m^r$ corresponding to the best neuron, i.e., $CR_m^r = CR_{i1}$. The first $F$ best neurons are then used at the next, $r + 1$, layer, and the training and selection of the neurons are repeated.

The value of $CR_m^r$ is step-by-step decreases while the number of layers increases, and the network grows up. Once, the value of $CR$ reaches to a minimal point and then starts to increase and we conclude that the network has been over-fitted. Because the minimum of $CR$ was reached at the previous layer, we stop the training algorithm and take a desirable network, which has been grown at the third layer.

## 4.2   A GMDH-Type Algorithm

The conventional GMDH-type algorithms perform an exhaustive search for candidate-neurons in each layer. The number of candidate neurons increases very fast with increasing the number $m$ of inputs as well as with the number $F$ of selected neurons. For the first and the next layers these numbers are $L_1 = \binom{m}{2}$ and $L_2 = \binom{F}{2}$. Below we describe the GMDH-type algorithm we developed and applied to induce the polynomial networks from data represented by $m > 70$ input features.

An idea behind this algorithm is to select the neurons one-by-one and then add them to the network with the calculated probabilities. For selecting the neurons it is used the exterior criterion described above.

In contrast to the exhaustive search, the algorithm randomly selects a pair of the neurons by using a "*roulette-wheel*" in which the wheel square is divided into $F$ sectors. The square of these sectors is proportional to the classification accuracy of the selected neurons on the training data. The neurons selected in the pair are then mated with a probability, which is proportional to their classification accuracy on the validating examples. Adding the new layer to the network, the algorithm attempts to improve the accuracy of the network for a given number of times.

```
X = [x₁, ..., xₘ]; % a pool of m input variables
k = 0; % the number of neurons in the network NN
% Train neurons with p inputs and calculate accuracy
p = 1;     % the number of inputs
```

```
for i = 1:m
 N1 = create-neuron(p, X(i));
 N1 = fit-weight(N1);
 A(i) = calc-accuracy(N1);
end
% Create new two-input neurons for gno attempts
p = 2;
for i = 1:gno
 pair = turn-roulette(p, A);
 N1 = create-neuron(p, X(pair));
 N1 = fit-weight(N1);
 ac = calc-accuracy(N1);
 % Selection and Addition
 if ac > max(A(pair))
   k := k + 1;
   NN(k) = add-new-neuron(N1);
   A(m + k) = ac;
 end
end
```

As a result, the variable NN contains description of the neural network. This network provides the best classification accuracy on the validating examples.

### 4.3   Classification of EEGs of Alzheimer and Healthy Patients

In our experiments we used the EEGs recorded from an Alzheimer and a healthy patient via the standard 19-channels C1, …, C19 during 8 seconds (Duke & Nayak 2002). Muscle artifacts were deleted from these data by an expert. We used the standard Fast Fourier Transform technique to calculate the spectral powers into four standard frequency bands: delta (0-4 Hz), theta (4-8 Hz), alpha (8-14 Hz) and beta (14-20 Hz).

As the spectral powers were calculated into ½ sec segments with ¼ sec overlapping, each EEG record consisted of 31 segments presented by 76 spectral features. The first 15 segments were used for training and the remaining 16 for testing, so the training and testing data consisted of 30 and 32 EEG segments, respectively.

Exploiting non-linear polynomial (5) and $F = 1$, our algorithm induced a polynomial network consisting of 4 input nodes and 3 neurons. In Figure 8 we depict this network. As we can see the induced classification rule is described by a set of 3 polynomials:

$y_1^{(1)} = 0.6965 + 0.3916 x_{11} + 0.2484 x_{69} - 0.2312 x_{11} x_{69}$,

$$y_1^{(2)} = 0.3863 + 0.5648 y_1^{(1)} + 0.5418 x_{73} - 0.4847 y_1^{(1)} x_{73},$$
$$y_1^{(3)} = 0.1914 + 0.7763 y_1^{(2)} + 0.2378 x_{76} - 0.2042 y_1^{(2)} x_{76}$$

where $x_{11}$ is delta in C11, $x_{69}$, $x_{73}$, and $x_{76}$ are beta in C12, C16, C19, respectively.

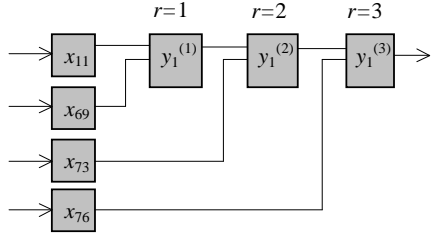

Figure 8: A polynomial network for classifying EEG of a Alzheimer and a healthy patient.

Note that medical experts can interpret these polynomials just as a weighted sum of two features, e.g., polynomial $y_1^{(1)}$ is interpreted as a weighted sum of features $x_{11}$ and $x_{69}$. The first two weights here show the significance of these features for the polynomial output and the third weight shows the power of interaction between these two features.

Having applied the standard neural network technique to these data, we found that a FNN, including 8 input nodes and 2 hidden neurons, provides the best classification accuracy. We also applied a conventional GMDH-type technique to these data. All three neural networks misclassified 1 testing segment, i.e., their testing error rate was 3.12%.

## 4.4    Recognition of EEG Artifacts

The EEGs used in our experiments were recorded from two sleeping newborns. These EEGs were presented by 72 spectral and statistical features as described in (Breidbach et al. 1998) calculated into 10-second segments. For training we used the EEG recorded from one newborn and for testing the EEG recorded from the other newborn. These EEGs consisted of 1347 and 808 examples in which an expert labeled 88 and 71 segments as artifacts, respectively.

For comparison, we used the standard neural network and the conventional GMDH techniques. We found out that the best FNN consisting of 10 hidden neurons has misclassified 3.84% out of testing examples. The GMDH-type network has been grown with an activation function (5) for $m = 72$ inputs and $F = 40$. We ran our algorithm with the same parameters which has induced a polynomial network consisting of seven input nodes and 11 neurons as depicted in Figure 9.

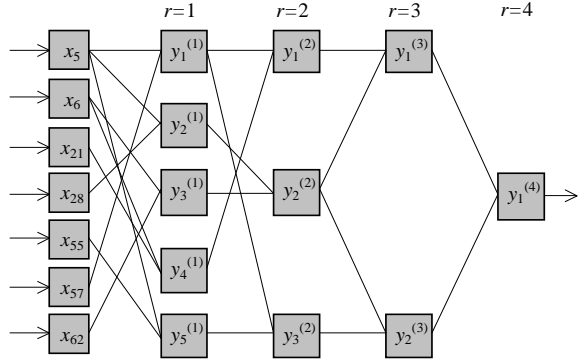

Figure 9: A polynomial network induced for recognizing EEG artifacts.

This polynomial network misclassified 3.47% out of testing examples. This network is described by the following set of 11 short-term polynomials:

$y_1^{(1)} = 0.9049 - 0.1707x_5 - 0.1616x_{57} + 0.0339x_5x_{57},$
$y_2^{(1)} = 0.9023 - 0.2128x_5 - 0.1389x_{28} + 0.0438x_5x_{28},$
$y_3^{(1)} = 0.9268 - 0.1828x_6 - 0.1195x_{62} + 0.0233x_6x_{62},$
$y_4^{(1)} = 0.9323 - 0.2057x_6 - 0.0461x_{21} + 0.0246x_6x_{21},$
$y_5^{(1)} = 0.9247 - 0.1822x_5 - 0.0951x_{55} + 0.0196x_5x_{55},$
$y_1^{(2)} = 0.0590 + 0.2810y_1^{(1)} + 0.3055y_4^{(1)} + 0.3670y_1^{(1)}y_4^{(1)},$
$y_2^{(2)} = 0.0225 + 0.4144y_2^{(1)} + 0.3812y_3^{(1)} + 0.1878y_2^{(1)}y_3^{(1)},$
$y_3^{(2)} = 0.0609 + 0.2917y_1^{(1)} + 0.2738y_5^{(1)} + 0.3880y_1^{(1)}y_5^{(1)},$
$y_1^{(3)} = 0.0551 + 0.3033y_1^{(2)} + 0.3896y_2^{(2)} + 0.2540y_1^{(2)}y_2^{(2)},$
$y_2^{(3)} = 0.0579 + 0.4058y_2^{(2)} + 0.2834y_3^{(2)} + 0.2549y_2^{(2)}y_3^{(2)},$
$y_1^{(4)} = -0.0400 + 0.6196y_1^{(3)} + 0.5702y_2^{(3)} - 0.1504y_1^{(3)}y_2^{(3)},$

where $x_5$ is the absolute power of subdelta in C4, $x_6$ is the absolute power of subdelta, $x_{21}$ is the real power of alpha, $x_{28}$ is the absolute power of beta1 in C3, $x_{55}$ is the absolute variance of theta in C4, $x_{57}$ the is absolute variance of subdelta and $x_{62}$ is the absolute variance of subdelta in C3.

Table 1 depicts the errors of the FNN, GMDH-type and polynomial neural networks (PNN) on the testing data. Note that both the FNN and the PNN were trained 100 times because their weights are initialized randomly. The conventional GMDH algorithm ran one time because it exploits the standard LSM technique of evaluating the synaptic weights.

Table 1: The classification errors of neural networks

|  | Error rate,% | | |
| --- | --- | --- | --- |
| *Data* | *FNN* | *GMDH* | *PNN* |
| Train (patient 1) | 2.00 | 2.06 | 2.23 |
| Test (patient 2) | 3.84 | 4.08 | 3.47 |

Observing the results listed in Table 1, we can conclude that the PNN trained by our method recognizes EEG artifacts slightly better than the FNN and GMDH-type network.

## 5. NEURAL NETWORK DECISION TREES

In this section we describe neural network decision tree techniques, which exploit the multivariate linear tests and the algorithms searching for relevant features. The linear tests are easily observable for medical experts. We also describe a new decision tree structure and an algorithm which is able to select the relevant features. This technique is shown to perform well on the large-scale clinical EEGs

### 5.1 Decision Trees

Decision tree (DT) methods have successfully been used for inducing multi-class concepts from real-world data represented by noisy features (Brodley & Utgoff 1995; Duda & Hart 2000; Quinlan 1993; Salzberg, Delcher, Fasman & Henderson 1998). Experts find that a DT is easy to observe by tracing the route from its entry point to the outcome one. This route may consist of the subsequence of questions which are useful for the classification and understandable for medical experts.

The conventional DTs consist of the *nodes* of two types. One is a *splitting* node containing a *test*, and other is a *leaf* node assigned to an appropriate class. A *branch* of the DT represents each possible outcome of the test. An example is presented to the *root* of the DT and follows the branches until the leaf node is reached. The name of the class at the leaf is the resulting classification.

The node can test one or more of the input variables. A DT is a *multivariate* or *oblique* one, if its nodes test more than one of the features. Multivariate DTs are in general much shorter than those which test a single variable. These DTs can test Threshold Logical Units (TLU) or *perceptrons* which perform a weighted sum of the input variables. Medical experts can interpret such tests just as a weighted sum of the questions, for example, is 0.4*ature for the test outcome.

To learn concepts presented by the numerical features Duda & Hart (2000), and Salzberg *et al*. (1998) have suggested multivariate DTs which allow classifying *linearly separable pattern*s. By definition such patterns are divided by linear tests. However using the algorithms (Brodley & Utgoff 1995; Frean 1992; Parekh, et al. 2000; Salzberg et al. 1998) such DTs can also learn to classify non-linearly separable examples.

In general, the DT algorithms require computational time that grows proportionally to the number of training examples, input features and classes. Nevertheless, the computational time, which is required to induce multi-class concepts from large-scale data sets, becomes overwhelming, especially, if the number of training examples is tens of thousands.

## 5.2   A Linear Machine

A Linear Machine (LM) is a set of *r linear discriminant* functions calculated to assign a training example to one of the $r \geq 2$ classes (Duda & Hart 2000). Each node of the LM tests a linear combination of $m$ input variables $x_1, x_2, \ldots, x_m$ and $x_0 \equiv 1$.

Let us introduce a $m$-input vector $\boldsymbol{x} = (x_0, x_1, \ldots, x_m)$ and a discriminant function $g(\boldsymbol{x})$. Then the linear test at the $j$th node has the following form:

$$g_j(\boldsymbol{x}) = \Sigma_i w_i^j x_i = \boldsymbol{w}^{jT}\boldsymbol{x} > 0, \ i = 0, \ldots, m, \ j = 1, \ldots, r, \tag{7}$$

where $w_0^j, \ldots, w_m^j$ are the real valued coefficients also known as a weight vector $\boldsymbol{w}^j$ of the $j$th TLU.

The LM assigns an example $\boldsymbol{x}$ to the $j$ class if and only if the output of the $j$th node is larger than the outputs of the other nodes:

$$g_j(\boldsymbol{x}) > g_k(\boldsymbol{x}), \ k \neq j = 1, \ldots, r. \tag{8}$$

This strategy of making a decision is known as Winner Take All (WTA).

During learning the LM, the weight vectors $\boldsymbol{w}^j$ and $\boldsymbol{w}^k$ of the discriminant functions $g_j$ and $g_k$ are updated on each example $\boldsymbol{x}$ that the LM misclassifies. A learning rule increases the weights $\boldsymbol{w}^j$, where $j$ is the class to which the example $\boldsymbol{x}$ actually belongs, and decreases the weights $\boldsymbol{w}^k$, where $k$ is the class to which the LM erroneously assigns the example $\boldsymbol{x}$. This is done using the following error correction rule:

$$\boldsymbol{w}^j := \boldsymbol{w}^j + c\boldsymbol{x}, \ \boldsymbol{w}^k := \boldsymbol{w}^k - c\boldsymbol{x}, \tag{9}$$

where $c > 0$ is a given amount of correction.

If the training examples are linearly separable, above procedure can yield a desirable LM giving maximal classification accuracy in a finite number of steps (Duda & Hart 2000). If the examples are non-linearly separable, this training procedure may not provide predictable classification accuracy. For this case the other training procedures have been suggested some of them we will discuss below.

### 5.3   A Pocket Algorithm

To learn the DT from data, which are non-linearly separable, Gallant (1993) had suggested a Pocket Algorithm. This algorithm seeks weights of multivariate tests that minimize the classification error. The Pocket Algorithm uses error correction rule (9) to update the weights $w^j$ and $w^k$ of the corresponding discriminant functions $g_j$ and $g_k$. The algorithm saves in the Pocket the best weight vectors $W^P$ that are seen during learning.

In addition, Gallant has suggested the "ratchet" modification of the Pocket Algorithm. The idea behind this algorithm is to replace of the weight $W^P$ by current $W$ only if the current LM has correctly classified more training examples than was achieved by $W^P$. The modified algorithm finds the optimal weights if sufficient training time is allowed.

To implement this idea, the algorithm cycles training the LM for the given number of epochs, $n_e$. For each epoch, the algorithm counts the current number of input series of correctly classified examples, $L$, and evaluates accuracy $A$ of the LM on the training set.

In correspondence to inequality (8), the LM assigns a training example ($x$, $q$) to the $j$th class, where $q$ is a class to which the example $x$ actually belongs. The LM training algorithm consists of the following steps:

```
W = init-weight();
[Wp, Lp, Ap] = set-pocket(W);
for i = 1:n % n is the number of training examples
  [x, q] = get-random(X);
  j = classify(x);
  if j ~= q
    Lp = 0;
    W(j) := W(j) + c*x;
    W(q) := W(q) - c*x;
  else
    if L > Lp
      A = calc-accuracy();
      if A > Ap
        % Update the pocket
        Wp = W;
        Lp = L;
        Ap = A;
      end
    end
  end
end
```

As searching time that the algorithm requires grows proportional to the numbers of the training examples as well as of the input variables and classes, the number of epochs must be large enough to achieve an acceptable classification accuracy. For example, in our case the number of the epochs is set to the number of the training examples. The best classification accuracy of the LM is achieved if *c* is equal to 1.

When the training examples are not linearly separable, the classification accuracy of LMs may be unpredictable large. There are two cases when the behavior of the LM is destabilized during learning. In the first case, a misclassified example is far from a *hyperplane* dividing the classes. In such a case the dividing hyperplane has to be substantially readjusted. Such relatively large adjustments destabilize the training procedure. In the second case, the misclassified example lies very close to the dividing hyperplane, and the weights do not converge.

To improve the convergence of the training algorithm, Grean (1992) has suggested a thermal procedure. This procedure decreases attention to the large errors by using the following correction

$$c = \beta/(\beta + k^2), \quad k = (w^j - w^i)^T x/(2x^T x) + \varepsilon,$$

where $\beta$ is a parameter initialized to 2, and $\varepsilon > 0.1$ is a given constant.

The parameter $\beta$ is adjustable during training as follows. First, the magnitudes of the weight vectors are summed. If sum value decreased for the current weight adjustment, but increased during the previous adjustment, the parameter $\beta$ is reduced: $\beta = a\beta - b$, where *a* and *b* are given constants.

This reducing $\beta$ enables the algorithm to spend more time for training the LM with small values of $\beta$ that are needed to refine the location of the dividing hyperplane. However, the experiments on the real-world classification problems show that the training time for the thermal procedure and the LM is comparable (Parekh et al. 2000).

## 5.4      Feature Selection Algorithms

In order to induce accurate and understandable DT models, we must eliminate the features that do not contribute to the classification accuracy of DT nodes. To eliminate irrelevant features, we use the Sequential Feature Selection (SFS) algorithms (Duda & Hart 2000; Galant 1993) based on a *greedy* heuristic, called also the *hill-climbing* strategy. The selection is performed while the DT nodes learn from data. This avoids over-fitting more effectively than the standard methods of feature pre-processing.

The SFS algorithm exploits a *bottom up search* method and starts to learn using one feature. Then it iteratively adds the new feature providing the largest improvement in the classification accuracy of the linear test. The algorithm continues to add the features until a specified stopping criterion is met. During this process the best linear test $T_b$ with the minimum number of the features is stored. In general, the SFS algorithm consists of the following steps.

```
p = 1; % the number of features in the test
% Test the unit-variant tests T
for i = 1:m
  T(i) = test(p, X);
end
Tb = find-best-test(T);
while stop-rule(Tb, p)
  p := p + 1;
  T1 = find-best-test(p, T);
  % Compare the accuracies of T1 and Tb
  if T1.A > Tb.A
    Tb = T1;
  end
end
```

The stopping rule is satisfied when all the features have been involved in the test. In this case $m + (m - 1) + \ldots + (m - k)$ linear tests have been made, where $k$ is the number of the steps. Clearly if the number of the features, $m$, as well as the number of the examples, $n$, is large, the computational time needed to terminate may be unacceptable.

To stop the search early and reduce the computational time, the following heuristic stopping criterion was suggested by Parekhet al. (2000). They found that if at any step the accuracy of the best test is decreased by more than 10%, then the chance of subsequently finding a better test with more features is slight.

However, the classification accuracy of the resulting linear test depends on the order in which the features have been included in the test. For the SFS algorithm, the order in which the features are added is determined by their contribution to the classification accuracy. As we know, the accuracy depends on the initial weights as well as on the sequence of the training examples selected randomly. For this reason the linear test can be non-optimal, i.e., the test can include more or fewer features needed for the best classification accuracy. The chance of selecting the non-optimal linear test is high, because the algorithm compares the tests that differ by one feature only.

## 5.5     **Induction of Neural Network Decision Trees**

The idea behind our DT induction algorithm is to individually train the test nodes and then group them in order to linearly approximate dividing hyperplanes. The DT test nodes, which are realized by TLUs, are individually trained to classify examples of two classes. For $r$ classes, therefore, it is needed to classify the $\binom{r}{2}$ variants of the training subsets and train the same number of TLUs.

Taking the trained TLUs dealing with one class, we can consider them as the hidden neurons of a neural network. The number of such networks is equal to the number of the classes, $r$. The contributions of these hidden neurons are summarized by the output TLU. Therefore each neural network makes linear approximation of the dividing hyperplane between classes.

Let us introduce a TLU $f_{i/j}$, performing a linear test (7), which learns to divide the examples of a pair of classes $\Omega_i$ and $\Omega_j$. If the training examples of these classes are linearly separable, then the output $y$ of the TLU is described as follows

$$y = f_{i/j}(\boldsymbol{x}) =\ \ 1, \ \forall\ \boldsymbol{x} \in \Omega_i, \tag{10}$$
$$y = f_{i/j}(\boldsymbol{x}) = -1, \ \forall\ \boldsymbol{x} \in \Omega_j.$$

Indeed, medical experts can find that the features dividing two classes are simpler to observe than those dividing the multiple classes for $r > 2$. Fortunately, when the number of classes does not exceed several tens, such a *pairwise* approach can be efficiently applied to a multi-class problem by transforming it to a set of simple binary classifiers.

Having introduced the linear tests, now we can illustrate the idea of our induction algorithm by a simple case of $r = 3$ classes. In Figure 10 we depict three classes $\Omega_1$, $\Omega_2$, and $\Omega_3$, which hardly overlap each other. For this simple case, we need to train the $\binom{r}{2} = 3$ TLUs. The lines in this Figure depict the hyperplanes $f_{1/2}$, $f_{1/3}$, and $f_{2/3}$ of the TLUs trained to divide the classes $\Omega_1$ and $\Omega_2$, $\Omega_1$ and $\Omega_3$, as well as $\Omega_2$ and $\Omega_3$.

In Figure 10 we depict also three new dividing hyperplanes $g_1$, $g_2$ and $g_3$. The first hyperplane $g_1$ is a superposition of the linear tests $f_{1/2}$ and $f_{1/3}$, i.e., $g_1 = f_{1/2} + f_{1/3}$. The linear tests $f_{1/2}$ and $f_{1/3}$ here are summarized with weights equal to 1, because both give us the positive outputs on the examples belonging to the class $\Omega_1$. Correspondingly, the second and third dividing hyperplanes are $g_2 = f_{2/3} - f_{1/2}$ and $g_3 = -f_{1/3} - f_{2/3}$.

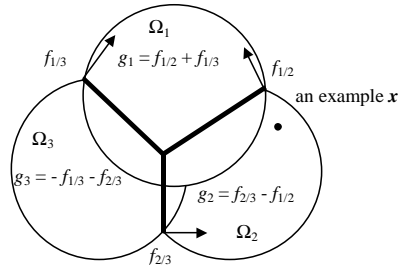

Figure 10: The approximation by the dividing hyperplanes $g_1$, $g_2$ and $g_3$.

We can see that an example $x$ belonging to class $\Omega_2$ causes the outputs of $g_1$, $g_2$ and $g_3$ to be equal to 0, 2, and –2, respectively:

$g_1(x) = f_{1/2}(x) + f_{1/3}(x) = 1 - 1 = 0,$
$g_2(x) = f_{2/3}(x) - f_{1/2}(x) = 1 + 1 = 2,$
$g_3(x) = -f_{1/3}(x) - f_{2/3}(x) = -1 - 1 = -2.$

We can see that among $g_1$, $g_2$, and $g_3$, the second output is largest, $g_2 = 2$. Then the DT, using the WTA strategy, correctly assigns the example $x$ to the class $\Omega_2$.

For this case, the dividing hyperplanes $g_1$, $g_2$, and $g_3$ were approximated by $r = 3$ feed-forward neural networks consisting of the $(r - 1) = 2$ hidden TLUs. In Figure 11, we depict these networks in which hidden neurons perform the linear tests $f_{1/2}$, $f_{1/3}$, and $f_{2/3}$, respectively. The hidden neurons are connected to the output neurons $g_1$, $g_2$ and $g_3$ with the weights equal to (+1, +1), (–1, +1) and (–1, –1), respectively.

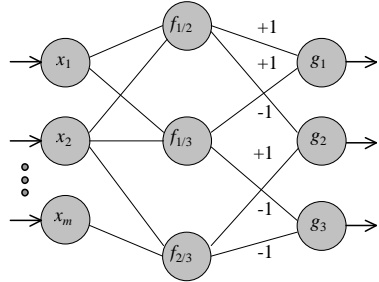

Figure 11: An example of the neural network decision tree for r = 3 classes.

In general for $r > 2$ classes, the neural network consists of $r(r - 1)/2$ hidden neurons $f_{1/2}, \ldots, f_{i/j}, \ldots, f_{r-1/r}$ and $r$ output neurons $g_1, \ldots, g_r$, where $i < j$, $j = 2, \ldots, r$. The output neuron $g_i$ is connected to $(r - 1)$ hidden neurons which are partitioned into two groups: the first group consists of the hidden neurons $f_{i/k}$ for which $k > i$, and the second group consists of the hidden neurons $f_{k/i}$ for which $k < i$. Finally it is set up the weights of output neurons: the output neuron $g_i$ are connected to the hidden neurons $f_{i/k}$ and $f_{k/i}$ with weights equal to +1 and -1.

As we see, each hidden neuron in the network learns to distinguish one class from another. The neurons learn independently of each other. However, the performance of the hidden neurons depends on the contribution of the input variables to the classification accuracy. For this reason below we discuss the DT induction algorithm which is able to select relevant features.

## 5.6    A Decision Tree Induction Algorithm

The feature selection algorithm that we discussed in Section 5.4 searches for new features, which cause the largest increases of the classification accuracy of the linear tests. We can see that, firstly, this algorithm compares the tests, which differ by one feature. Secondly the algorithm uses the greedy heuristic to select the new feature, which provides the largest increasing in the accuracy of the current test.

In our experiments we found that the comparison between the linear tests, which differ by more than one feature, increases the chance of accepting those tests, which improve the classification accuracy of the DT. We found also that in real-world classification problems represented by noisy data the greedy heuristic often finds a local minimum of classification error. To enlarge the chance of escaping from the local minima, we can evaluate the cross-validation classification error of linear tests. Using these heuristics, we developed a DT induction algorithm shown below:

```
X = [x_1, x_2,…, x_m];
% Test the unit-variant tests U
for i = 1:m
  U(i) = train-test(X(i));
  C(i) = calc-accuracy(U(i));
end
% Create the pools P and F
P = C/max(C);
[P, F] = sort-descend(P);
Tb = [];  % initialize
Ab = 0;
for k = 1:attempt-no
  T = [];
  i = 0;
  feature-no = 0,
  % Search for a candidate-test T
  while stop-rule(T, i)
```

```
    i := i + 1;
    if P(i) > rand(1)  % wheel of roulette
      T1 = [T X(F(i))]; % the features of test
      T1 = train-test(T1);
      A1 = calc-accuracy(T1);
      if A1 > A
        T := T1;
        A := A1;
        feature-no := feature-no + 1;
      end
    end
  end
  % Replace the best test Tb
  if A > Ab
    Ab := A;
    Tb := T;
  end
end
```

To search out a best multivariate test, this algorithm exploits an evolving strategy: it starts to train the single-variable tests including one feature $x_i$, $i = 1, \ldots, m$. Then, it calculates a probability $p_i$ that is proportional to the accuracy of the *i*th test.

The calculated values of the probabilities are arranged in decreasing order, i.e., $p_{i1} \geq p_{i2} \geq \ldots \geq p_{im}$, and then they are placed in a pool $P$. Likewise the features $x_{i1}, x_{i2}, \ldots, x_{im}$ are placed in a pool $F$.

In the next steps the algorithm sets an empty array [] to the test $T$ and 1 to the index. Then it attempts to add the feature $x_i$ to $T$. If this occurs with the calculated probability $p_i$, then a candidate-test $T_1$ is formed. The weights of this test are fitted to the training data, and then the classification accuracy $A_1$ of the test on the validation test is calculated.

If the accuracy $A_1$ becomes higher than the accuracy $A$ of the current test $T$, then $T$ is replaced by the candidate-test $T_1$. The number of features used in the new linear test $T_1$ increases by one.

The algorithm is repeated until a stopping criterion is met. This criterion is met in two cases: first, if the linear test $T$ includes the given number $N_f$ of the input variables, $m$, or second, if all the features have been tested.

Note that the algorithm compares the linear tests $T$ and $T_b$ which may differ by several features. This increases the chance of searching out a best linear test.

To increase the chance of locating the best solution, the linear tests are trained by the given number $N_a$ attempts, each time with the different sequence of features. As a result, a unique set of the features is formed in the test $T_b$. Using these features, the linear test classifies the training examples with the best classification accuracy $A_b$.

For fitting the DT linear tests, we used 2/3 of the training examples and evaluated the classification accuracy on all the training data. We varied the number of attempts $N_a$ from 5 to 25.

## 5.7 Learning a Multi-Class Concept from the EEGs

The above DT algorithm has been applied for learning a multi-class concept from the clinical EEG recordings. The EEGs were recorded from 65 sleeping patients via the standard EEG electrodes C3 and C4. These patients were healthy newborns aged between 35 and 51 weeks. So, the desired concept must distinguish the EEG recordings between these $r = 16$ age groups (classes).

Following Breidbach et al. (1998), the raw EEGs were segmented and transformed to 72 spectral and statistical features. Some of these features were redundant or irrelevant to the classification problem.

For training and testing the DT, we used 39399 and 19670 EEG segments respectively. For a given $r = 16$ classes, the DT included the $r(r-1)/2 = 120$ simple binary classifiers. The training errors of these classifiers varied between 0 and 15%, see Figure 12(a).

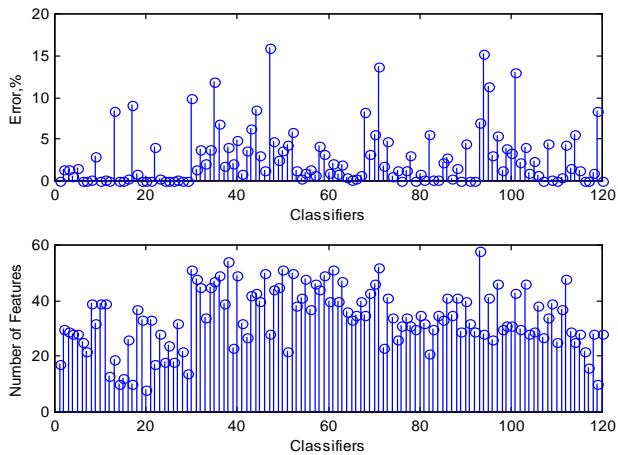

Figure 12: The training errors (a) and the number of features (b) for 120 binary classifiers.

Note that the trained classifiers use different sets of the features (input variables). The number of these features varies from 7 to 58, see Figure 12(b).

The trained neural network DT correctly classified the 80.8% of the training and 80.1% of the testing examples. Summing all the segments belonging to one EEG recording, the trained DT correctly classified 89.2% and 87.7% of the 65 EEG recordings on the training and testing examples, respectively.

In Figure 13 we depict the distributions of the classified testing segments over all 16 classes for two patients belonging to the second and third age groups, respectively. Observing these distributions, we can give a probabilistic interpretation of the decisions. For example, we can decide that the patients belong to the second and third age groups with probabilities 0.92 and 0.58, respectively.

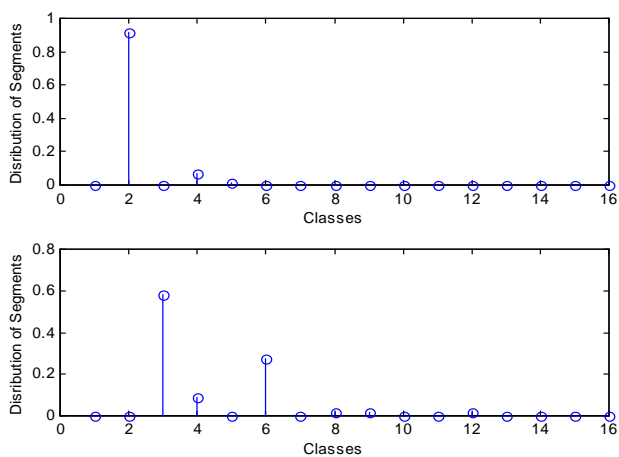

Figure 13: The distribution of the classified testing segments for two patients.

We compared this DT technique with some data mining techniques on the same EEG data. First, we induced the LM described in section 5. Second, we trained the feed-forward neural networks by the standard back-propagation algorithm. The structures of the neural networks included from 8 to 20 input nodes and up to 20 hidden neurons. Third, we trained independently $r = 16$ binary classifiers to distinguish one class against others. Fourth, we trained a binary decision tree consisting of $r - 1 = 15$ linear classifies. However, in our experiments these standard techniques could not achieve desirable classification accuracy.

## 6.   A RULE EXTRACTION TECHNIQUE

In some cases the classification models learnt from data by the neural-networks can be represented as decision tree rules (Avilo Garcez et al 2001; Sethi & Yoo 1997). However in general this technique cannot guarantee that a resultant decision tree was not trapped in a local solution (Kovalerchuk & Vityaev 2000). Taking this in account below we describe our technique developed to induce the decision tree rules in *ad hoc* manner.

The idea behind our method is to project an original classification problem into an input space in which most of the training examples become separable. The dimensionality of such an input space can be significantly less than that of the original space. The neural-network techniques described in sections 3 and 4 are well suitable for this role because they outperform the standard neural networks.

Indeed, removing the misclassified examples from the training data and eliminating the noise and irrelevant features from the original feature set, we can significantly simplify the class boundaries and the solution of the classification problem. A decision tree induced from such data can be well suited in order to classify new observations well.

To describe our technique, let us assume that the polynomial neural network performs enough well on the testing data and define the training subsets X0 and X1 consisting of $n_0$ and $n_1$ examples which have been correctly assigned by this network to the classes 0 and 1. These examples are represented in the new space of features, V, whose dimensionality is now equal to $m$. Then the DT induction algorithm can be described as follows.

```
T = [];   % a decision tree T = ∅
V = 1:m;  % a pool V of features
find-node(X0, X1, V);
```

The procedure `find-node` is invoked with parameters X0, X1, and V. This procedure adds a new node to the decision tree T and then recursively calls itself as follows:

```
m = number-of-features(V);
% Search a threshold q_i and an outcome p_i
for i = 1 to m do
  [q_i, p_i]= search-threshold-and-outcome();
end
[v_1, e_1]= find-feature-dividing X0 and X1;
f_1 = create-new-test();
T = [T, f_1]; % add new test to T;
% Calculate the outputs Y0 and Y1
Y0 = f_1(X0);
Y1 = f_1(X1);
```

```
V = remove-feature(v₁);
if V not empty
 % Find the examples A0, A10, A1 and A01:
 A0  = find(Y0 == 0);
 A01 = find(Y0 == 1);   % the errors of 0
 A1  = find(Y1 == 1);
 A10 = find(Y1 == 0);   % the errors of 1
 if A10 not empty
   find-node(X0(A0, V), X1(A10, V), V);
 end
 if A01 not empty
   find-node(X0(A01, V), X1(A1, V), V);
 end
end
```

We used this algorithm to induce a decision tree for recognizing the artifacts in the clinical EEGs. First we learnt the polynomial network described in section.4.4 from the training data which have been originally presented by 72 features. Then we removed from these data all 30 misclassified examples and used the 7 discovered features to present the data in the new input space.

To induce a DT from the new data the above algorithm has been applied. This algorithm has induced a simple DT which exploits only one variable $x_6$, the absolute power of subdelta summed over channels C3 and C4, as depicted in Figure 14.

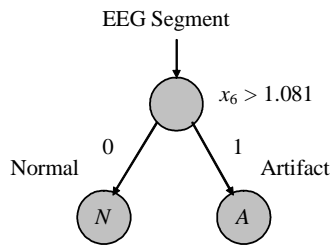

Figure 14: A decision tree rule for classifying the normal EEG segments and artifacts.

Surprisingly, this decision tree has misclassified 24 testing examples, whilst an original polynomial network did 28. More experimental results providing the evidence can be found in the paper devoted to the artifact recognition in the clinical EEGs (Schetinin & Schult 2004).

So we can see that EEG experts can easily understand and interpret this decision tree as follows: an EEG segment is the artifact, if the value of absolute power of subdelta, $x_6$, is more than 1.081, otherwise, it is normal segment.

## 7.  CONCLUSION

Standard neural networks can learn classification rules from real-world data well, however such classification models cannot be comprehensible for experts. The classification models become to be more understandable if they are represented in a visual form. To achieve such a representation, data mining techniques, based on a strategy of searching for a trade-off between complexity and accuracy of classification rules, are commonly used. In contrast to this strategy, the methods described in this chapter allow experts to present classification models in visual form and keep their classification error down.

We applied the standard and our neural network techniques to clinical EEG data to extract classification models which EEG-experts could easily present visually. On testing data these models performed slightly better than the standard feed-forward and GMDH-type networks. Thus we conclude that our neural-network techniques can be successfully used for visual data mining clinical EEGs.

## 8.  ACKNOWLEDGMENTS

The work has been supported by the University of Jena (Germany) and particularly by the University of Exeter (UK) under EPSRC Grant GR/R24357/01. The authors are personally grateful to Frank Pasemann for fruitful discussions, Joachim Frenzel and Burkhart Scheidt for clinical EEG recordings we used in our experiments, to Richard Everson and Jonathan Fieldsend for useful comments.

## 9.  EXERCISE AND PROBLEMS

1.  Assume a fully connected neural network consisting of 5 input nodes, 3 hidden and 2 output neurons. What minimal number of examples is required to train this network by back-propagation? Why user does need to preset a structure of neural network? What is changed in the neural network if the user applied PCA and determined 2 principle components?

2.  Suppose a continue eXclusive OR (XOR) problem is described as follows

$y = 1$, if $x_1 x_2 > 0$, and $y = 0$, if $x_1 x_2 \leq 0$,

where $y$ are a target output, and $x_1 \in [-1, 1]$, $x_2 \in [-1, 1]$ are the input variables.

If the user uses a fully connected neural network, what structure has to be preset for this problem?

3. Assume an evolved cascade neural network which consists of 3 hidden neurons and 1 output neuron. How many examples are required to train this network? How many neurons are required to solve XOR problem?

4. Suppose the GMDH-type neural network exploiting a transfer polynomial

$$y = g(v_1, v_2) = w_0 + w_1v_1 + w_2v_2 + w_3v_1v_2 + w_4v_1^2 + w_5v_2^2,$$

where $v_1$ and $v_2$ are the input variables and $w_0, \ldots, w_5$ are the coefficients.

What minimal number of examples is required to train this network? When GMDH-type neural networks out-perform fully connected neural networks and *vice versa*?

5. When and why multivariate decision trees outperform decision trees which test single variables? Regarding to the above XOR problem, which of these techniques is better?

6. Assume a 4-class problem. How many neurons are required to learn linear machine? What structure of a neural network based on pairwise classification should be preset for this case?

7. When and why multi-class systems based on pairwise classification outperform the standard neural networks and decision trees?